\documentclass{svproc}
\usepackage{url}
\usepackage{graphicx}

\usepackage{amsmath}
\usepackage[ruled,vlined]{algorithm2e}
\DeclareMathOperator*{\argmax}{arg\,max}
\begin{document}
\mainmatter              
\title{DeepSwarm: Optimising Convolutional Neural Networks using Swarm Intelligence}
\titlerunning{DeepSwarm: Optimising CNNs using Swarm Intelligence}  
%
\author{Edvinas Byla \and Wei Pang}
\authorrunning{Edvinas Byla \and Wei Pang} 
%
\tocauthor{Edvinas Byla and Wei Pang}
\institute{
Department of Computing Science, University \\of Aberdeen, Aberdeen AB24 3UE, UK\\
\email{e.byla.15@aberdeen.ac.uk}, \\
\email{pang.wei@abdn.ac.uk}
}

\maketitle              

\begin{abstract}

In this paper we propose DeepSwarm, a novel neural architecture search (NAS) method based on Swarm Intelligence principles. At its core DeepSwarm uses Ant Colony Optimization (ACO) to generate ant population which uses the pheromone information to collectively search for the best neural architecture. Furthermore, by using local and global pheromone update rules our method ensures the balance between exploitation and exploration. On top of this, to make our method more efficient we combine progressive neural architecture search with weight reusability. Furthermore, due to the nature of ACO our method can incorporate heuristic information which can further speed up the search process. After systematic and extensive evaluation, we discover that on three different datasets (MNIST, Fashion-MNIST, and CIFAR-10) when compared to existing systems our proposed method demonstrates competitive performance. Finally, we open source DeepSwarm\footnote{\url{https://github.com/Pattio/DeepSwarm}} as a NAS library and hope it can be used by more deep learning researchers and practitioners.

\end{abstract}
\section{Introduction \label{sec:introduction}}

In recent years it has become increasingly challenging for human engineers to manually design deep neural architectures for specific tasks. This is mainly due to the following two facts: (1) modern deep neural architectures tend to be very complex with a lot of layers and hyperparameters; (2) one architecture might perform well on one dataset or on one type of problems but poorly on others. These two factors have resulted in a boom of research that tries to develop methods that can automate the design of neural architectures, the so-called neural architecture search \cite{wistuba2019survey}.

In this paper we propose a novel neural architecture search method based on Swarm Intelligence (SI). To start with, we focus on Convolutional Neural Networks (CNN) \cite{lecun2015deep}, one of the most commonly used deep neural architectures. To discover new CNN architectures our method uses Ant Colony Optimization (ACO) \cite{dorigo_phd}. The motivation for using SI for NAS is due to the fact that SI possesses many appealing properties that could be helpful when dealing with NAS problems. This includes fault tolerance, decentralisation, scalability and ability to share and combine the knowledge, just to name a few. In particular, ACO has few distinct characteristics that make it naturally fit into the NAS domain: ACO is good at solving discrete problems which can be represented as graphs and it can easily adapt to dynamic environment (changing graph). Another significant motivating factor to use SI is the fact that the majority of its methods have not been explored in the context of NAS. 

The novel contributions of this research are summarised as follows: 
\begin{itemize}
    \item We show that ACO can be used to effectively optimise CNNs.

    \item We use heuristic information when performing NAS based on ACO.
    
    \item We dynamically change the graph size and progressively search for the architectures when performing NAS based on ACO.
\end{itemize}

The rest of the paper is organised as follows: Section \ref{sec:related-work} presents related work; Section \ref{sec:deepswarm} introduces our proposed method; Section \ref{sec:experiments} presents the evaluation of our method; and Section \ref{sec:conclusion} concludes the paper and explores possible future directions.

\section{Related Work \label{sec:related-work}}

Neural Architecture Search (NAS) is an automated process that aims to discover the best performing neural network architectures for a specific problem. Even though NAS research goes back as far as three decades \cite{miller1989designing}, it has attracted new attention in recent years with the rapid development of deep learning, significant improvements in hardware, and growing interest of the machine learning community. Furthermore, even with this renewed interest from many deep learning researchers and practitioners it still seems that most of the existing NAS research predominantly focuses on using Evolutionary Algorithms \cite{real2017large,suganuma2017genetic,miikkulainen2019evolving}, Bayesian Optimisation \cite{domhan2015speeding,jin2018efficient}, and Reinforcement Learning \cite{zoph2016neural,baker2016designing,zoph2018learning}. However, considering most of these approaches require huge amounts of computational resources, some new work which tries to reduce the computational costs have emerged \cite{negrinho2017deeparchitect,elsken2017simple,pham2018efficient}. For example, in \cite{pham2018efficient} the authors proposed to use large computational graph which stores all the weights, and they reported that sharing these weights among child models could be 1000 times less computationally expensive than standard NAS approaches.

To the best of our knowledge, ACO was first applied to NAS problem in 2014 \cite{salama2014novel}, and in their work ACO was used to optimise feed-forward neural networks. Furthermore, in their work the authors discovered that reusing the weights of the best solution can further improve the performance of their method. In 2015 ACO was used to optimise the structure of deep recurrent neural networks \cite{desell2015evolving}, where the authors try to address the problem of predicting general aviation flight data. The authors reported that using ACO they could achieve better prediction performance for airspeed, altitude, and pitch compared with the previous best published results. Finally, in more recent work \cite{elsaid2018using}, ACO was used to optimise long short-term memory recurrent neural networks, and they achieved an increase in prediction
accuracy, while also reducing the number of trainable weights by 55\%.

It is noted that another relevant work to our research is the Progressive Neural Architecture Search (PNAS) approach \cite{liu2018progressive}: similar to PNAS, the system proposed in this paper explores enormous CNN search space by using small incremental steps. In \cite{liu2018progressive} the authors concluded that PNAS can achieve the same level of performance as the previous NAS approach \cite{zoph2018learning} while being 8 times faster in terms of the required total computational time.

\section{DeepSwarm \label{sec:deepswarm}}

In this section we first present the details of the proposed DeepSwarm, and then we give the overall workflow.  

As mentioned before, DeepSwarm search for new architectures in the order of increasing complexity similar to PNAS. At the beginning of a NAS task, DeepSwarm creates an internal graph which contains only the input node. Then a specified number of ants are generated. Next, one by one each ant is placed on the input node. After being placed on the input node each ant uses the Ant Colony System (ACS) \cite{dorigo_ant_colony_system} selection rule to select one of the available nodes in the next layer of CNN, and the ACS selection rule is as follows:
\begin{equation} \label{eq:pheromone_transition}
  s=\begin{cases}
    $$ \argmax\limits_{u \in J_k(r)}  \{ [\tau(r,u)] \cdot [\eta(r,u)]^\beta \}$$, & \text{if $q \leq q_0$ \quad (exploitation)}.\\
    S, & \text{otherwise \quad (biased exploration)},
  \end{cases}
\end{equation}

In the above $\tau(r,u)$ denotes the pheromone amount on the edge that goes from node $r$ to node $u$ and $\eta(r,u)$ denotes the heuristic value associated with the edge going from node $r$ to node $u$. Furthermore, $J_k(r)$ denotes a set of nodes that are available to visit from node $r$. The value of $q$ is a random number uniformly distributed over $[0\dots1]$. Parameters $q_0 \in (0,1]$ and $\beta \in (0,\inf)$ control the algorithm's greediness and the relative importance of heuristic information. Finally, S is a random variable selected according to the probabilistic distribution defined by Equation (\ref{eq:pheromone_distribution}):

\begin{equation}
  p_k(r,s)=\begin{cases}
    $$ \frac{[\tau(r,s)] \cdot [\eta(r,s)]^\beta}{\sum\limits_{u \in J_k(r)} [\tau(r,u)] \cdot [\eta(r,u)]^\beta}$$, & \text{if $s\in J_k(r)$}.\\
    0, & \text{otherwise}.
  \end{cases}
  \label{eq:pheromone_distribution}
\end{equation}

Once a node is selected the system checks if this node already exists in the graph at the depth of the selection. If this node is a new one which does not exist in the graph, it is added to the graph as a neighbour node to the previous node (i.e., the node where the ant was before the selection) so the subsequent ants can exploit the pheromone information. After an ant selects a particular node it also performs the same selection rule as defined by Equations (\ref{eq:pheromone_transition}) and (\ref{eq:pheromone_distribution}) to select the attributes of that node (i.e. filter size, kernel size). When the selection is completed the node is added to the ant's path. Once an ant reaches the current maximum allowed depth, its path is transformed into a neural network architecture which then gets evaluated. Furthermore, after an ant finishes a walk it performs ACS local pheromone update as defined by Equation (\ref{eq:pheromone_local_update}) for each edge it has used:
\begin{equation}
\tau(r,s) \longleftarrow (1-\rho) \cdot \tau(r,s) + \rho \cdot \tau_0
\label{eq:pheromone_local_update}
\end{equation}

In the above, parameter $\rho$ denotes the pheromone decay factor and parameter $\tau_0$ is the initial pheromone value. This local update rule decays pheromone values so the other ants can be encouraged to explore other paths. After all ants are evaluated the best ant is found (the ant which found the architecture with the highest accuracy). This best ant then performs the ACS global pheromone update as defined by Equation (\ref{eq:pheromone_global_update}), which increases the pheromone values for the edges found in the best path. 

\begin{equation}
\tau(r,s) \longleftarrow (1-\alpha) \cdot \tau(r,s) + \alpha \cdot \Delta \tau(r,s),
\label{eq:pheromone_global_update}
\end{equation}

\noindent where

\begin{equation}
\Delta \tau(r,s) = 
    \begin{cases}
        $$C_{gb}$$, & \text{if $(r,s) \in$ global-best-tour}. \\
        0, & \text{otherwise}.
    \end{cases}
    \label{eq:delta_tau}
\end{equation}

\begin{figure}[ht]
 \begin{center}
  \includegraphics[width=0.73\textwidth]{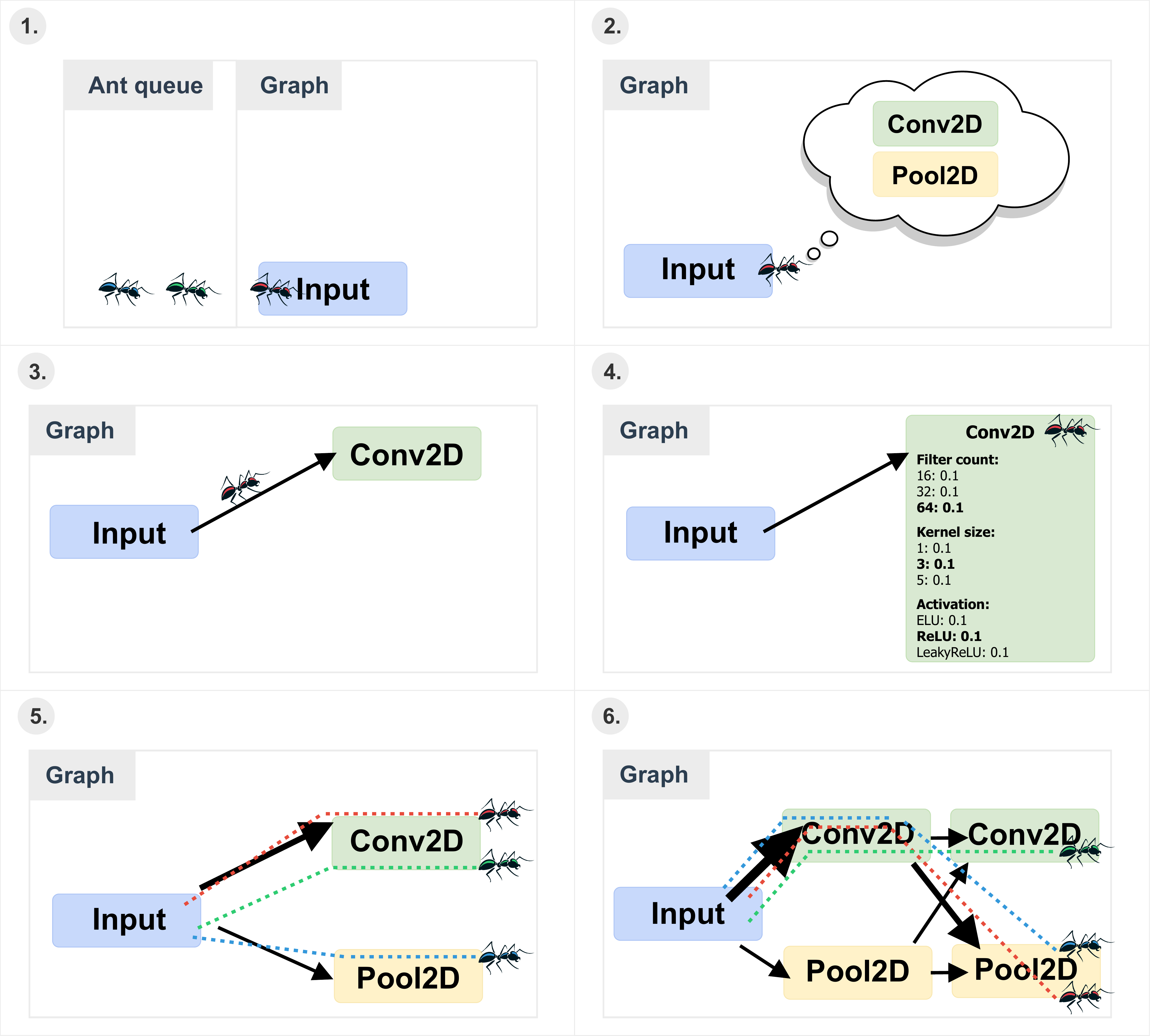}
  \caption{An overview of the NAS process of DeepSwarm. (1) The ant is placed on the input node. (2) The ant checks what transitions are available. (3) The ant uses the ACS selection rule to choose the next node. (4) After choosing the next node the ant selects the node's attributes. (5) After all ants finished their tour the pheromone is updated. (6) The maximum allowed depth is increased and the new ant population is generated. \textbf{Note}: Arrow thickness indicates the pheromone amount, meaning that thicker arrows have more pheromone. \label{fig:deepswarm-low-level-workflow-design}}
 \end{center}
\end{figure}

Here parameter $\alpha$ controls pheromone evaporation and its range is $(0, 1)$. $C_{gb}$ is the cost of the global best tour (the best model accuracy). After the graph's current maximum allowed depth is increased, a new population of ants is generated. This cycle is repeated until the maximum depth (specified by the user) is reached. An illustrative example of NAS performed by DeepSwarm can be seen in Fig. \ref{fig:deepswarm-low-level-workflow-design}, and the pseudocode is given in Algorithm \ref{algo:nas_pseudocode}.

\begin{algorithm}[ht!]
  \DontPrintSemicolon
  \SetKwFunction{FMain}{search}
  \SetKwFunction{FSecond}{generate\_ants}
  \SetKwFunction{FThird}{generate\_path}
  \SetKwFunction{FFourth}{aco\_select}
  \SetKwProg{Fn}{Function}{:}{}
  \Fn{\FMain{}}{
    graph = Graph() \tcp{build graph containing only the input node}
 \While{graph.current\_depth $<$ max\_depth}{
  ants = generate\_ants()\;
  best\_ant = find\_best(ants)\;
  graph.global\_pheromone\_update(best\_ant)\;
  graph.increase\_depth()\;
 }
   \Return best\_ant
  }
  \;
  \SetKwProg{Pn}{Function}{:}{\KwRet}
  \Pn{\FSecond{}}{
        ants = []\;

  \For{$i=0$ \KwTo $ant\_count$}{
  ant = Ant()\;
  ant.path = generate\_path()\;
  ant.evaluate()\;
  ants.append(ant)\;
  graph.local\_pheromone\_update(ant)\;
 }
 \Return ants
  }
  \;
  \SetKwProg{Pn}{Function}{:}{\KwRet}
  \Pn{\FThird{}}{
   current\_node = graph.input\_node\;
   path = [current\_node]\;
   \For{$i=0$ \KwTo $current\_max\_depth$}{
    \If{$current\_node.neighbours \longleftarrow \emptyset$}{
        \textbf{break}\;
    }
    current\_node = aco\_select\_rule(current\_node.neighbours)\;
    path.append(current\_node)
   }
   completed\_path = complete\_path(path) \tcp{completes the path if needed}
   \Return path
  }
  \;
  \SetKwProg{Pn}{Function}{:}{\KwRet}
  \Pn{\FFourth{neighbours}}{
  \ForEach{neighbour $\in$ neighbours}{%
      probability = neighbour.pheromone $\times$ neighbour.heuristic\;
      probabilities $\gets$ probability\;
      denominator += probability
    }
  \If{random.uniform(0, 1) $\leq$ greediness }{
      max\_index = probabilities.index(max(probabilities))\;
      \Return neighbours[max\_index]
    }
    probabilities = probabilities $/$ denominator \;
    neighbour\_index = wheel\_selection(probabilities)\;
    \Return neighbours[neighbour\_index]
  }
 \caption{DeepSwarm \label{algo:nas_pseudocode}}
\end{algorithm}

We point out several interesting outcomes of using ACO as a search strategy as follows: (1) weight reusability is straightforward to implement: we find the longest common sub-path in the graph and reuse the best weights from that sub-path, (2) the search space can be explored progressively as ants can adapt to the dynamic environment (when we expand the graph from depth $n$ to $n+1$ we do not lose the information which was gathered up to depth $n+1$), and (3) because ACO uses domain-specific heuristics (Equations (\ref{eq:pheromone_transition}) and (\ref{eq:pheromone_distribution})) domain experts can easily provide their own knowledge to speed up the search further.

\section{Experiments \label{sec:experiments}}

For the experimental design, three different datasets were chosen: (1) MNIST \cite{lecun1998mnist}, (2) Fashion-MNIST \cite{fashion-mnist}, and (3) CIFAR-10 \cite{cifar-10}. Each of these three datasets is quite different from the others and requires different CNN architectures to achieve the best results. As a result the combination of them is a good way to test the algorithm's robustness and performance. In order to evaluate our proposed method the baselines taken from \cite{jin2018efficient} were used. All of our tests were carried out in the Google Colab environment (1x Tesla K80 GPU) \cite{colab} using a MacBook Pro (Early 2015 model) to interact with this environment. Note that even though in \cite{jin2018efficient} they ran each method only for 12 hours, they used NVIDIA GeForce GTX 1080 Ti GPU, which according to a few benchmarks is approximately 2-3 times faster than our selected Tesla K80 GPU. This is the reason why we are not going to constrain our runs to 12 hours. 

\subsection{Evaluation Procedure \label{subsec:evaluation-procedure}}

When evaluating the system the following procedure was followed: (1) create a new Google Colab instance, (2) import the source code of the library, (3) split the training set 90-10 to training and validation sets, (4) run the algorithm until the max depth is reached, (5) take the best found network, (6) for CIFAR-10 dataset apply standard data augmentation (random horizontal flips, rotation and scaling) to the training data, (7) train the best found network for additional 50 epochs on the augmented data, (8) load the weights which showed the best performance on the validation set during those 50 epochs, and (9) evaluate the network with these best weights on the testing data.

\subsection{Ant Count \label{subsec:ant-count}}

\begin{figure}[ht]
\centering
\begin{minipage}{0.48\textwidth}
  \centering
  \includegraphics[width=1\textwidth]{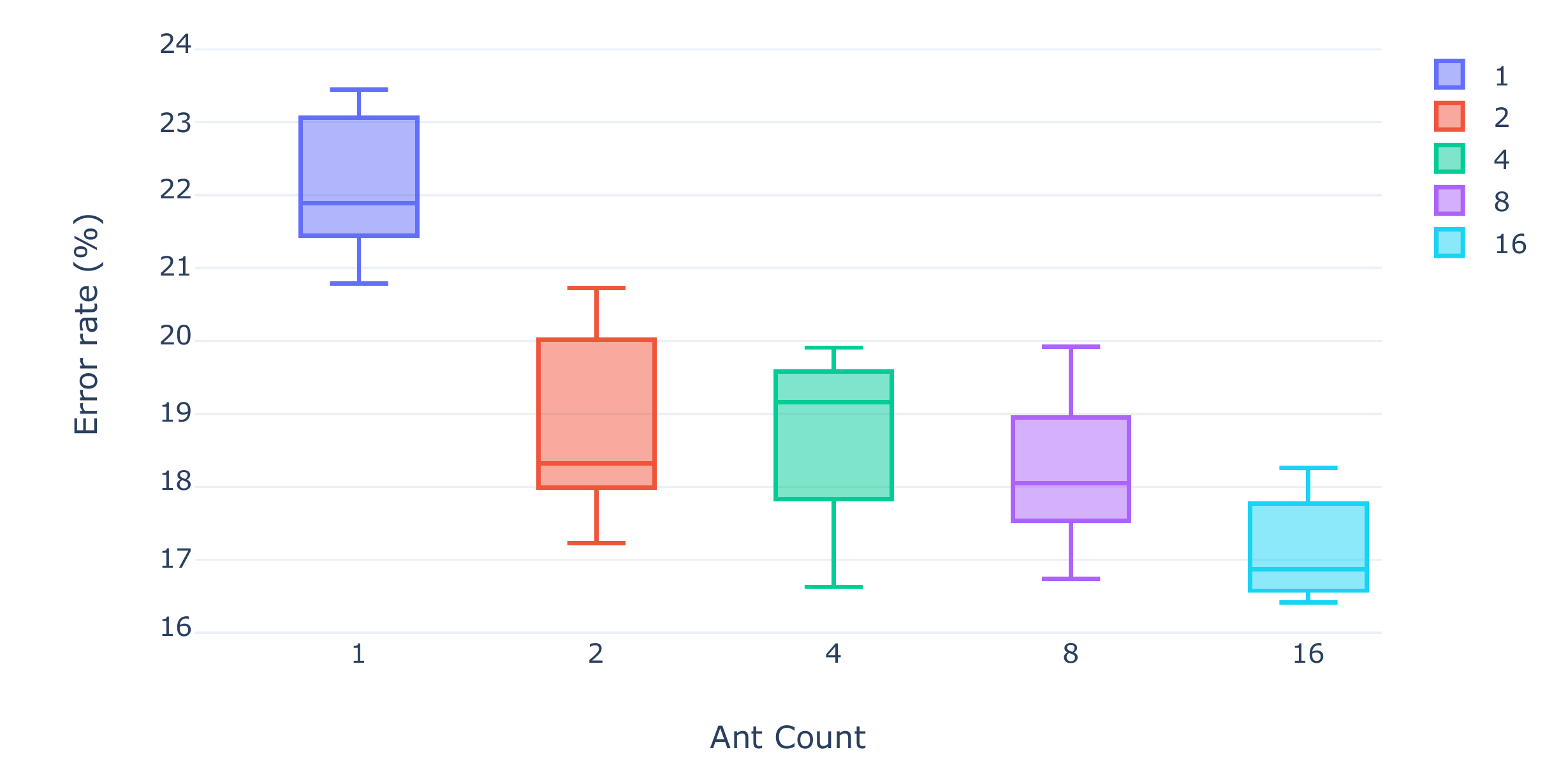}
  \caption{The error rate on the CIFAR-10 dataset before the final training across five separate trials. \label{fig:ant-count-before-train}}
\end{minipage}%
\hfill
\begin{minipage}{0.48\textwidth}
  \centering
  \includegraphics[width=1\textwidth]{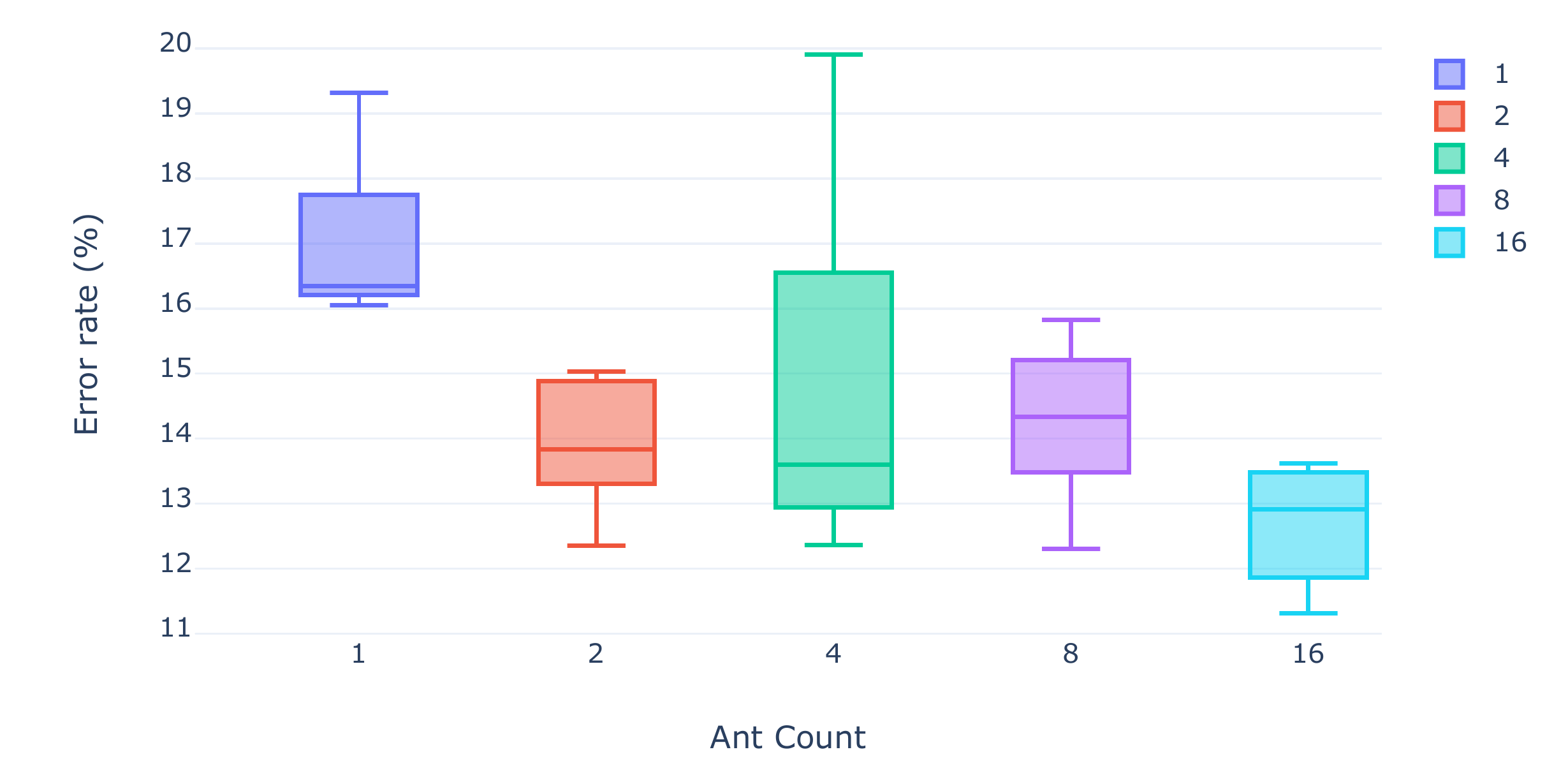}
  \caption{The error rate on the CIFAR-10 dataset after the final training across five separate trials. \label{fig:ant-count-after-train}}
\end{minipage}
\end{figure}

\begin{figure}[ht]
 \begin{center}
  \includegraphics[width=0.54\textwidth]{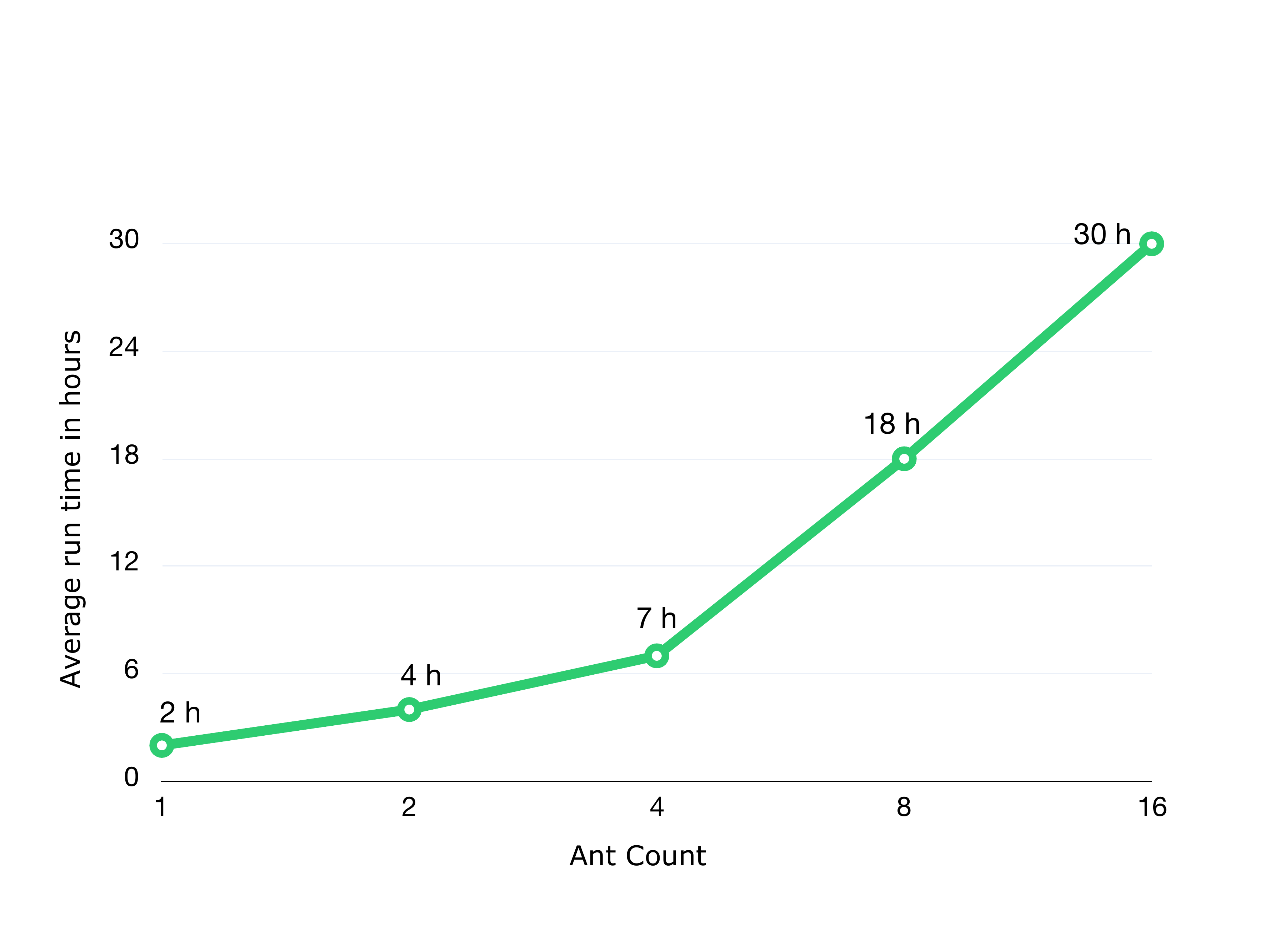}
  \caption{The average run time (across five different trials) in hours for different ant counts on the CIFAR-10 dataset. \label{fig:ant-run-time}}
 \end{center}
\end{figure}

The ant count  (the number of ants used during search) is one of the most important hyperparameters in DeepSwarm. This is because it is a trade-off between the performance of the final model and the run-time of the algorithm. In order to find a good trade-off, we ran multiple tests by exponentially increasing the ant count. Furthermore, we split the results into two parts: before and after the final training. Before the final training is a part where DeepSwarm finds potentially the best model and after the final training is the part where the best found model is trained for an additional 50 epochs on augmented data. The reason for this choice is that the results before the final training can reflect the real implications that the ant count has on the error rate, whereas the results after the final training can show how the ant count can affect the generalisation. This follows from the fact that before the final training the models are trained on the same data, whereas during the final training the models are trained on the augmented data which can show how well they can learn. The results before the final training are presented in Fig. \ref{fig:ant-count-before-train}, the results after the final training can be seen in Fig. \ref{fig:ant-count-after-train}, and the run time is shown in Fig. \ref{fig:ant-run-time}.

Looking at the results one can see that changing the ant count from 1 to 2 had a significant impact on the error rate. This finding was to be expected because when only one ant exists both exploration and exploitation must suffer. The exploration suffering is associated with the fact that the ant can only explore one architecture per depth, meaning that only a small subset of available architectures will be explored. The exploitation degradation occurs because at each depth acquired knowledge scales only linearly, for example, at depth 3 the ant will only know about 2 other architectures. Furthermore, having only one ant will result in rather greedy behaviour where the same ant will explore the same sub-tree in the graph and will only rarely explore the parallel sub-trees. We further noticed that even though doubling the ant count almost doubles the run time, it will not always result in drastically improved performance. For example, when we increased the ant count from 4 to 8 ants the run time increased from 7 hours to 18 hours, while the average error rate decreased only by 0.13\%. The most drastic changes in the error rate happened when the ant count was changed from 1 to 2 (3.11\% decrease) and from 8 to 16 (2.1\% decrease). However, due to the computational restrictions we did not test ant counts beyond 16 which means that there might be even bigger performance improvements when going beyond 16 ants.

\subsection{Greediness \label{subsec:grediness}}

\begin{figure}[h]
\centering
\begin{minipage}{0.48\textwidth}
  \centering
  \includegraphics[width=1\textwidth]{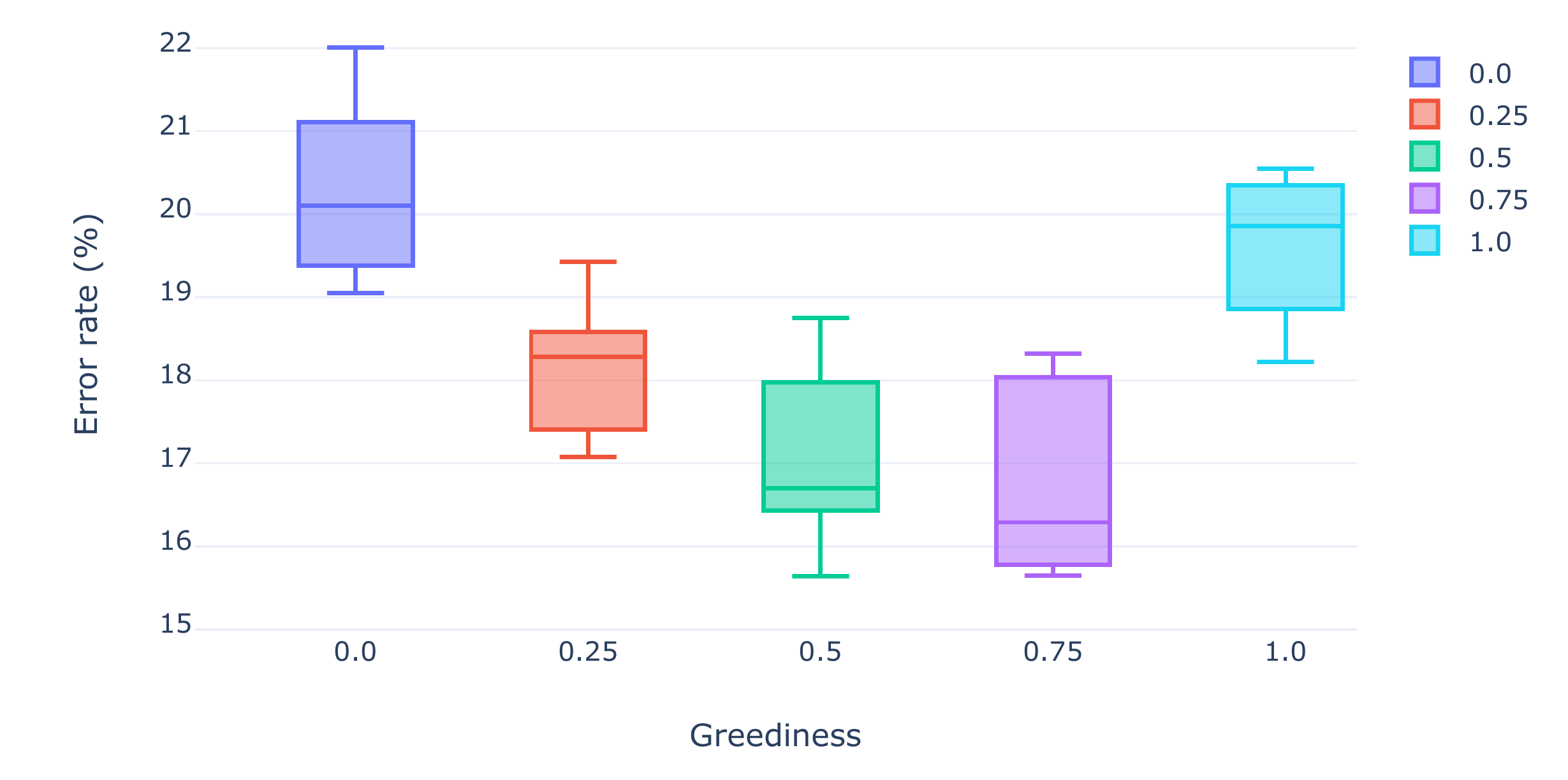}
  \caption{The error rate on the CIFAR-10 dataset before the final training across five separate trials. \label{fig:greediness-before-train}}
\end{minipage}%
\hfill
\begin{minipage}{0.48\textwidth}
  \centering
  \includegraphics[width=1\textwidth]{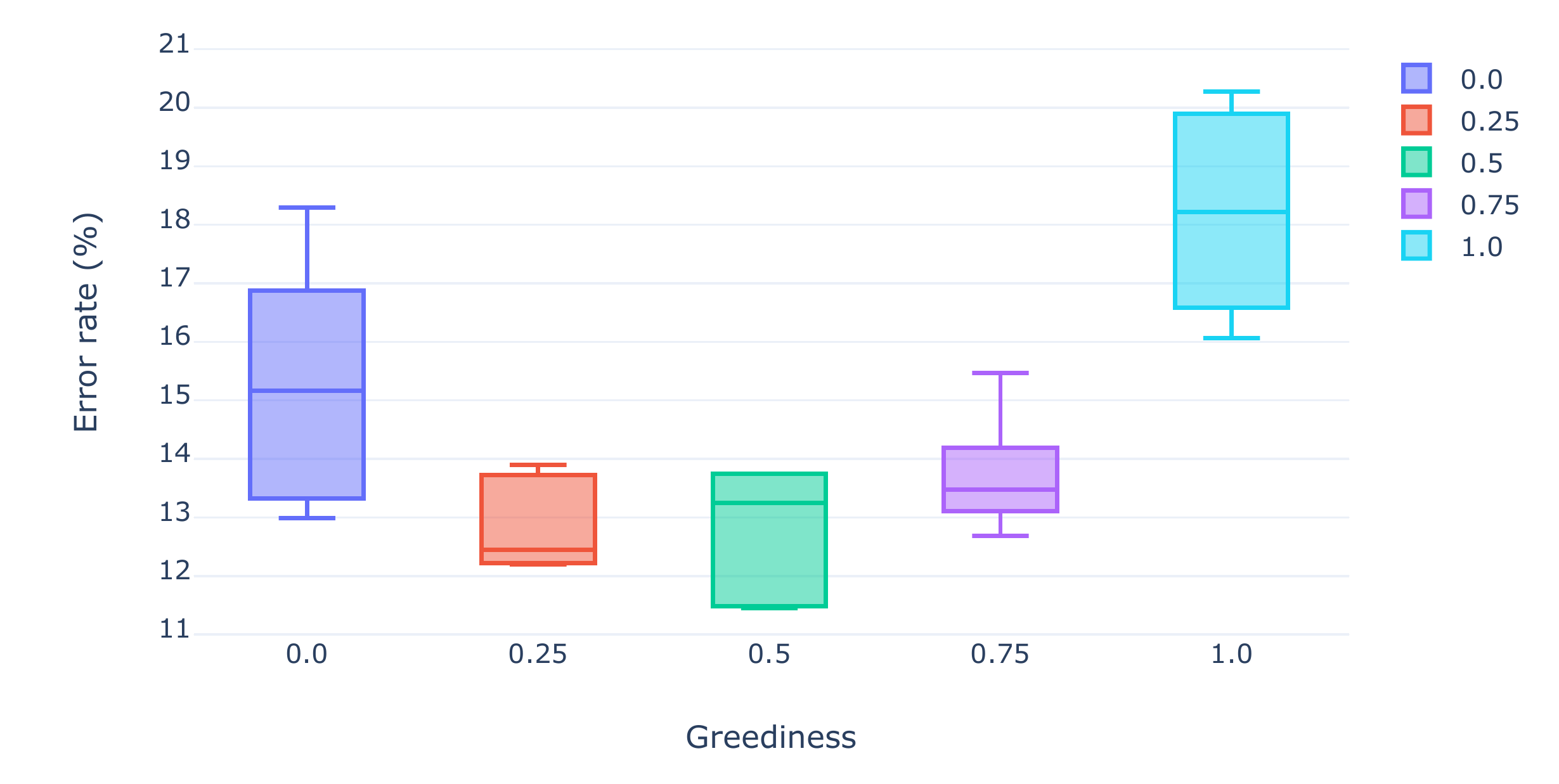}
  \caption{The error rate on the CIFAR-10 dataset after the final training across five separate trials. \label{fig:greediness-after-train}}
\end{minipage}
\end{figure}

Another important hyperparameter of DeepSwarm is greediness. As mentioned in Section \ref{sec:deepswarm}, the greediness is used in Equation (\ref{eq:pheromone_transition}) to decide how greedy each ant should be. As greediness can be defined in the range from 0.0 to 1.0, we test the greediness with its value increases from 0 to 1 at a step size of 0.25. Furthermore, similarly to the ant count, the results were divided into before and after the final training. The results before the final training are shown in Fig. \ref{fig:greediness-before-train} and the results after the final training are shown in Fig. \ref{fig:greediness-after-train}. 

Looking at the results it seems that when selecting the greediness for the algorithm one should never go to extremes as this will most likely result in poor performance. The more general insight we gathered from the results was that selecting the greediness values which were close towards the middle (0.5) resulted in the best performance. The reason why the extremely greedy ants perform poorly is as follows: at the beginning of the search they base their search purely on the heuristic information and then, once the pheromone is laid on the graph, all of them will reuse the same path, therefore generating the same architecture. Furthermore, the local pheromone update rule will not help here because once the pheromone evaporates these greedy ants will use the same heuristics which will result in the same paths being chosen again. In contrast, the ants with no greediness will always base each of their decisions only on the wheel selection without exploiting the gathered information (as the first part in Equation \ref{eq:pheromone_transition} is always skipped) and because during the path generation an ant needs to make a lot of these decisions (choosing the next node and each attribute), a substantial part of them will be random, which will result in a poor performance. Another interesting observation was that the greedy models tend to generalise worse than the less greedy ones. For example, the average error rate difference before the final training between 0.25 greediness and 0.75 greediness was 1.32\% (18.12\% and 16.80\% respectively), but after the final training, the difference was -0.83\% (12.89\% and 13.72\% respectively). Furthermore, we noticed that the greediness had some impact on the average network depth, for example, the best architectures which were found using no greediness, were on average five layers deeper than the ones which were found using 1.0 greediness. As a result of that, these less greedy architectures had more regularisation and feature extraction. We believe that this could be the reason why these less greedy architectures were generalising better during the final training.

\subsection{Accuracy \label{subsec:accuracy}}

In order to compare the performance of DeepSwarm with that of other methods, we report the average and best performance achieved during the five separate runs on three different datasets. These results are shown in Table \ref{table:accuracy-table}. From these results we can see that on the MNIST dataset from all of the methods DeepSwarm showed the best performance. When compared with the straightforward methods (random and grid search \cite{bergstra2012random}) DeepSwarm showed a significantly lower error rate (1.79\%, 1.68\% versus 0.46\%). On the Fashion-MNIST dataset, DeepSwarm achieved the lowest error rate and once again proved to be superior to the straightforward methods which had almost a two times bigger error rate (11.36\%, 10.28\% versus 6.75\%). Finally, on the CIFAR-10 dataset, even though DeepSwarm managed to find the architecture with the lowest error rate (11.31\%), on average its performance was not as good as some other methods. Overall on all of the three datasets, DeepSwarm still produced very competitive and promising results. To see the best architectures discovered by DeepSwarm please refer to Appendix \ref{Appendix}.

\begin{table}[ht]
\centering
     \begin{tabular}{c c c c} 
     \hline
     Method & MNIST & Fashion-MNIST & CIFAR-10 \\
     \hline
     RANDOM & 1.79\% & 11.36\% & 16.86\%\\
     GRID & 1.68\% & 10.28\% & 17.17\% \\
     \hline
     SPMT & 1.36\% & 9.62\% & 14.68\%\\
     SMAC & 1.43\% & 10.87\% & 15.04\%\\
     \hline
     SEAS & 1.07\% & 8.05\% & 12.43\% \\
     NASBOT & N\char`\/A & N\char`\/A & 12.30\% \\
     \hline
     AutoKeras BFS & 1.56\% & 9.13\% & 13.84\% \\
     AutoKeras BO & 1.83\%  & 7.99\% & 12.90\% \\
     AutoKeras BFS & 0.55\% & 7.42\% & \textbf{11.44}\%\\
     \hline
     DeepSwarm Average & \textbf{0.46\%} & \textbf{6.75\%} & 12.70\% \\
     DeepSwarm Best & 0.39\% & 6.44\% & 11.31\% \\
     \hline
    \end{tabular}
\caption{The error rates on the CIFAR-10 dataset. \label{table:accuracy-table}}
\end{table}

\subsection{Discussion \label{discussion}}

Even though there exists a NAS approach developed by Google Brain \cite{zoph2018learning} which can achieve better results than DeepSwarm on the CIFAR-10 dataset, we think that it would be not fair to compare our work with theirs for the following reasons: (1) they used 400 GPUs (also their GPUs were much more powerful than the one used in our experiments) for 4 days, (2) they used skip and add connections which are not implemented into DeepSwarm yet. We also point out that as they did not open source their code, it is not easy for us to test their approach in our environment to compare the performance difference. Nevertheless, based on the results seen in Section \ref{subsec:accuracy} DeepSwarm proved to be a competitive approach against already existing NAS methods. However, there is still some work that needs to be done in order to further improve DeepSwarm. We think that the two main components that can be added in the future are skip and add nodes. Adding these two components would allow DeepSwarm to search for more complex architectures which in turn could substantially improve the overall learning performance. Finally, we list the main advantages of DeepSwarm compared with other existing NAS systems as follows:

\begin{itemize}
    \item DeepSwarm offers competitive performance. As shown in Section \ref{subsec:accuracy}, on all 3 datasets DeepSwarm can achieve comparable or better results than the other NAS systems. 
    
    \item DeepSwarm can look for diverse structures. DeepSwarm does not enforce a specific structure, which allows it to find novel and interesting architectures.
    
    \item DeepSwarm can offer fast search. As mentioned earlier, DeepSwarm is built to search for architectures progressively and has a mechanism to reuse the old weights which boosts its performance. 
    
    \item DeepSwarm allows the users to provide heuristic information which can further speed up the search process.
    
    \item DeepSwarm is easy to use. To start the neural architecture search a user just needs to write a few lines of code (see detailed instructions on DeepSwarm's GitHub page).
    
    \item DeepSwarm is easy to be further developed and extended. As we open source DeepSwarm and share it with the wider machine learning community, other researchers can further develop and extend DeepSwarm.
\end{itemize}

\section{Conclusion and Future Work \label{sec:conclusion}}

In this paper we presented DeepSwarm and demonstrated that Swarm Intelligence can be used to effectively tackle NAS problems. After evaluating DeepSwarm we discovered that when compared to other similar methods it can show competitive performance. Furthermore, we open source DeepSwarm\footnote{\url{https://github.com/Pattio/DeepSwarm}} and share it with the community, and we hope more people will benefit from it and further develop it.

The main contribution of this work is to show that ACO can be used to effectively search for optimal CNN architectures. Our second contribution is to demonstrate that domain expert knowledge can be successfully incorporated into ACO based NAS. The final contribution of this work is to show that progressive architecture search approach can be applied to ACO based NAS methods.

For future work we propose to explore the following directions: (1) implement skip and add connections which would allow ants to look for more complex architectures, (2) try to use ACO to perform cell based search (similar to \cite{zoph2018learning}) rather than the full architecture search, (3) compare conventional search method with the progressive search when ACO is applied to NAS problem, and (4) explore ACO in other deep learning contexts i.e. find which neurons to drop in the dropout layer.

%
%

\bibliography{mybib}

\clearpage
\appendix
\section{Appendix \label{Appendix}} 

\begin{figure}[h!]
 \begin{center}
  \includegraphics[height=0.85\textheight]{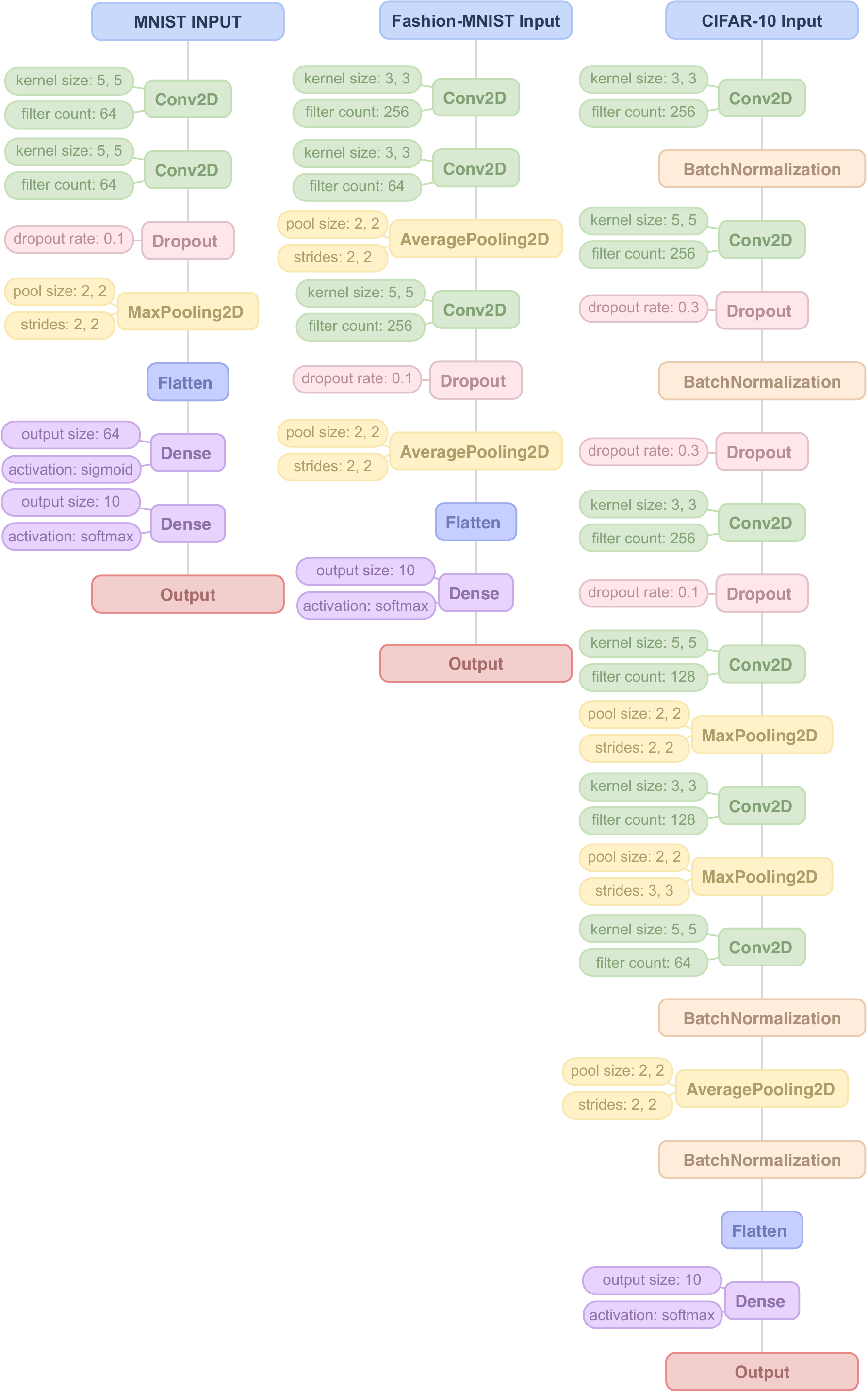}
  \caption{The best architectures discovered by DeepSwarm. \label{fig:best-architectures}}
 \end{center}
\end{figure}

\end{document}